\documentclass[letterpaper, 10 pt, conference]{ieeeconf}
\IEEEoverridecommandlockouts
\usepackage{graphicx}
\usepackage{amsmath}
\usepackage{amssymb}
\usepackage{subcaption}
\usepackage{booktabs}

\title{\LARGE \bf
MfNeuPAN: Proactive End-to-End Navigation in Dynamic Environments via Direct Multi-Frame Point Constraints
}

\author{
Yiwen Ying$^{1, 2}$, Hanjing Ye$^{1, 2}$, Senzi Luo$^{1, 2}$, Luyao Liu$^{1, 2}$, Yu Zhan$^{1, 2}$, Li He$^{1, 2}$ and Hong Zhang$^{1, 2 *}$ \textit{Fellow IEEE}
\thanks{$^{1}$All authors are with the Department of Electronic and Electrical Engineering, Southern University of Science and Technology, Shenzhen, China. {\tt\small \{yingyw2022, yehj2022, luosz2024, liuly2023, 12232151\}@mail.sustech.edu.cn, \{hel, hzhang\}@sustech.edu.cn}}
\thanks{$^{2}$All authors are with Shenzhen Key Laboratory of Robotics and Computer Vision, Shenzhen, China.}
\thanks{*Corresponding author.}
\thanks{This work was supported in part by the Shenzhen Science and Technology Program (No. SGDX20240115111759002), in part by the Meituan Academy of Robotics Shenzhen, in part by the SUSTech High-Level Special Funds (No. G03034K003), in part by the National Natural Science Foundation of China under Grant No. 62173096, and in part by the Pearl River Talent Recruitment Program under Grant No. 2019QN01X761.}
}

\begin{document}

\maketitle
\thispagestyle{empty}
\pagestyle{empty}

\begin{abstract}
Obstacle avoidance in complex and dynamic environments is a critical challenge for real-time robot navigation. Model-based and learning-based methods often fail in highly dynamic scenarios because traditional methods assume a static environment and cannot adapt to real-time changes, while learning-based methods rely on single-frame observations for motion constraint estimation, limiting their adaptability. To overcome these limitations, this paper proposes a novel framework that leverages multi-frame point constraints, including current and future frames predicted by a dedicated module, to enable proactive end-to-end navigation. By incorporating a prediction module that forecasts the future path of moving obstacles based on multi-frame observations, our method allows the robot to proactively anticipate and avoid potential dangers. This proactive planning capability significantly enhances navigation robustness and efficiency in unknown dynamic environments. Simulations and real-world experiments validate the effectiveness of our approach.
\end{abstract}

\section{Introduction}
\label{sec:introduction}
Robust navigation with obstacle avoidance is critical for deploying autonomous robots in complex, dynamic environments. It requires efficient and adaptive planning systems; however, its development is significantly complicated by the unpredictable nature of real-world conditions.

Traditional model-based methods, such as optimization-based techniques \cite{Zhang2020Optimization} and sampling-based planners \cite{Salzman2016Asymptotically}, have been extensively utilized in structured environments. These methods, which rely on predefined models of the environment and obstacles, offer accurate results in static settings. However, they often fail to address the complexity and uncertainty of real-world scenarios, where obstacles are highly dynamic and the environment changes rapidly. This is because many traditional methods, while not necessarily assuming a completely static environment, still struggle with the high computational cost associated with frequent model updates and the inability to quickly re-plan paths in response to sudden changes in the environment. As a result, they may not be well-suited for highly dynamic scenarios.

Learning-based methods, which leverage machine learning algorithms to adapt to new environments by learning from large datasets \cite{Fan2020Distributed}, have emerged as a viable alternative \cite{Devo2020Towards}. They handle complex and dynamic environments more effectively by identifying patterns and features from data, rather than relying on explicit models \cite{Qureshi2020Motion}. However, contemporary approaches like DRL-VO \cite{Xie2023DRL}, NavRL \cite{Xu2025NavRL}, and NeuPAN \cite{Han2025NeuPAN} still face challenges in dynamic scenarios. These methods are sensitive to environmental variations, resulting in poor performance when real-world conditions differ from the training data. They also exhibit limited generalization ability to unseen scenarios, especially those that differ significantly from the environments used during training. While works like DRL-VO rely on pattern recognition for rapid responses, this can result in error propagation. NeuPAN attempts to mitigate this by estimating motion planning constraints directly from the point cloud in an end-to-end manner, but it primarily uses points from the current frame for estimation, resulting in ineffective performance when dealing with dynamic obstacles that require predictive or anticipatory planning.

\begin{figure}
    \centering
    \includegraphics[width=\linewidth]{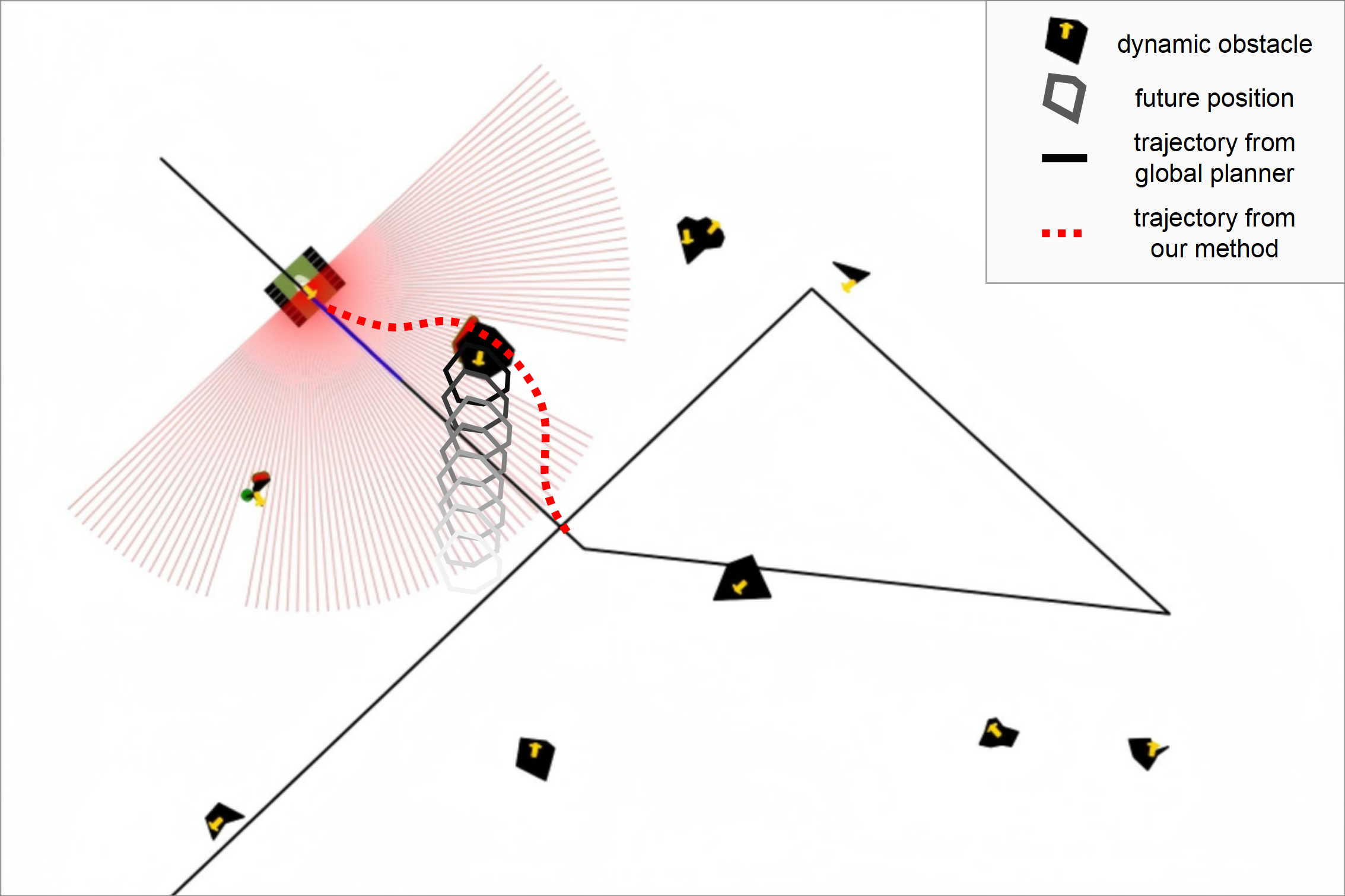}
    \caption{\textbf{Illustration of our method working in a dynamic environment.} When a dynamic obstacle crosses a predetermined trajectory, the robot avoids it in the opposite direction of the predetermined trajectory, demonstrating its proactive planning ability.}
    \label{fig:concept}
\end{figure}

In summary, both traditional model-based methods and contemporary learning-based methods struggle to effectively handle the high computational demands, sensitivity to environmental variations, limited generalization ability, and the need for anticipatory planning in dynamic environments. These limitations highlight the need for more advanced and adaptive approaches that can better address the challenges posed by complex, dynamic environments.

To address these challenges, this paper introduces MfNeuPAN, a framework based on NeuPAN \cite{Han2025NeuPAN} that incorporates additional proactive planning capabilities for handling dynamic objects. Instead of simply using single-frame point positions to estimate the motion planning constraints, we predict dynamic objects' positions in future frames and aggregate multi-frame points to estimate the constraints. In this way, planners can proactively plan to avoid dynamic objects.

The remainder of this paper is organized as follows. Section \ref{sec:related_work} reviews related work in robot navigation and obstacle avoidance. Section \ref{sec:method} details the proposed framework. Section \ref{sec:experiments} presents experimental results and ablation studies validating the framework. Section \ref{sec:conclusion} concludes the paper.

\section{Related Work}
\label{sec:related_work}
\subsection{Dynamic Environment Perception}

Accurate perception is essential for reliable navigation in dynamic environments. Early efforts leveraged event cameras for their high temporal resolution, enabling detection of fast-moving objects \cite{Falanga2020Dynamic}. However, their limitations in constructing static environment maps restricted their utility in full navigation pipelines.

To overcome these challenges, image-based methods such as background subtraction and optical flow were adopted \cite{Lin2020Robust}, but they often lacked robustness and incurred high computational costs in complex scenes \cite{Pokle2019Deep}. Many relied on YOLO-based detection \cite{redmon2016you}, which required obstacle pattern labeling—an unnecessary step for basic avoidance and a potential source of error.

Point cloud processing techniques have significantly advanced this field. DBSCAN clustering \cite{Ester1996Density} effectively segmented obstacles, while Kalman filters \cite{Chen2022Real} enabled motion state estimation. Nevertheless, real-time processing of dense point clouds remained challenging on mobile platforms \cite{Arpino2021Robot}. In parallel, learning-based methods, particularly deep neural networks, have shown prospects in improving segmentation accuracy \cite{Mersch2022Receding}. Yet, their computational demands limited real-time applications, especially in resource-constrained settings \cite{Tai2016ADeep}. Recently, Lu et al. \cite{Lu2024FAPP} introduced FAPP, using novel distance metrics for efficient point cloud segmentation in dynamic environments. This method allowed for the dynamic classification of obstacles into static, moving, or unknown categories, marking a breakthrough in real-time perception for cluttered environments.

A fundamental dichotomy exists between modular and end-to-end approaches. Modular systems decompose navigation into sequential components, which are intuitive and explainable, but suffer from error propagation across modules, where inaccuracies in perception degrade planning performance \cite{Fan2020Distributed}. By contrast, end-to-end methods directly map sensor inputs to control outputs, eliminating intermediate representations \cite{Devo2020Towards}. However, these methods often lack interpretability and require extensive retraining for new platforms \cite{Qureshi2020Motion}. NeuPAN \cite{Han2025NeuPAN} bridges this gap through its end-to-end model-based learning framework, which directly processes raw point clouds to generate physically interpretable motions while maintaining mathematical guarantees. In this paper, we adopt a similar framework as NeuPAN, but introduce a novel multi-frame prediction module to improve its adaptability in dynamic environments.

\subsection{Dynamic Obstacle Avoidance}

Dynamic obstacle avoidance has evolved notably, with key contributions shaping current methodologies. In the ROS \cite{Quigley2009ROS} system, the dynamic window approach \cite{Fox1997Dynamic} served as a local planner for obstacle avoidance. An influential framework was introduced by Fiorini and Shiller \cite{Fiorini1998Motion}, who proposed the concept of velocity obstacle (VO). VO’s simplicity and efficiency made it widely adopted in dynamic settings. Building upon this, Van den Berg et al. \cite{VanDenBerg2011Reciprocal} extended this to reciprocal velocity obstacles (RVO), incorporating multi-agent interactions for more realistic avoidance.

Further advancements came with the integration of Model Predictive Control (MPC) \cite{GARCIA1989335}, allowing trajectory optimization over finite time horizons \cite{Brito2019Model}. Lin et al. \cite{Lin2020Robust} applied MPC with visual input for micro-UAVs, enhancing 3D obstacle avoidance. However, these approaches often assume constant velocity or acceleration \cite{Lin2020Robust, Xu2022DPMPC}, limiting their adaptability to the irregular motions common in real-world environments, especially over multiple frames or time steps.

Deep reinforcement learning (DRL) introduced a new paradigm \cite{Xie2023DRL}, showing potential but facing issues such as reward sensitivity \cite{Xu2025NavRL}, sim-to-real transfer difficulties \cite{Liu2020Robot}, and poor interpretability in safety-critical contexts.

Model-based and learning-based planners each have trade-offs. Classical model-based techniques include graph-search methods, sampling-based planners, and optimization-based approaches. MPC-based optimization methods generate feasible trajectories under dynamic constraints \cite{Rosmann2017Kinodynamic} but struggle with computational complexity when scaling to dense environments \cite{Han2023RDA}. Learning-based methods, such as those using reinforcement learning \cite{Qureshi2020Motion, Wang2022Adaptive}, bypass explicit modeling by learning navigation policies from data, yet require careful reward design and lack transparency \cite{Li2021MPC}. Hybrid approaches like MPC-MPNet \cite{Li2021MPC} combine neural networks with classical planning but still suffer in unstructured environments. Furthermore, learning-based methods generally cannot ensure safety, limiting their standalone use in industrial applications where safety is critical.

NeuPAN \cite{Han2025NeuPAN} represents a significant advance by formulating navigation as an end-to-end optimization problem solved via a Plug-and-Play Proximal Alternating Minimization Network. NeuPAN outperforms previous methods (e.g., OBCA \cite{Zhang2020Optimization}, RDA \cite{Han2023RDA}, and Hybrid-RL \cite{Fan2020Distributed}) in accuracy, generalization, and latency. However, its original formulation relies on single-frame observations, limiting its performance in highly dynamic scenarios, by which the motion of moving obstacles cannot be adequately described. Our work extends NeuPAN by integrating multi-frame point constraints, and thereby accurately characterizing object motion to support robust navigation in dense, cluttered environments with moving obstacles.

\section{Method}
\label{sec:method}
\begin{figure*}[htbp]
    \centering
    \includegraphics[width=0.99\textwidth]{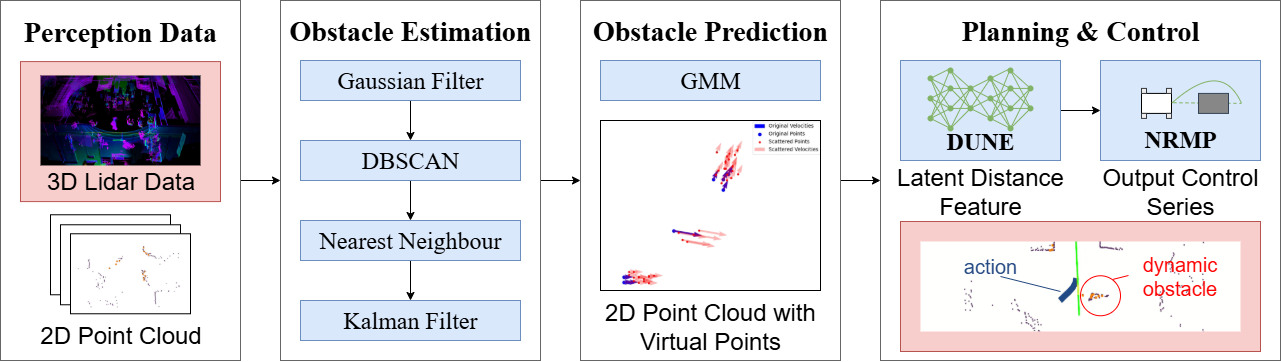}
    \caption{\textbf{System overview.} The obstacle estimation module processes point clouds to estimate obstacle motion states. Prediction forecasts obstacle trajectories. Planning and control generates control series.}
    \label{fig:system}
\end{figure*}

Our end-to-end navigation framework processes multi-frame lidar scans, each comprising 3D points relative to the robot’s current pose. It uses point positions from the latest 10 frames and predicted positions for the next 20 steps to capture obstacle dynamics and improve motion constraint estimation. The framework includes DBSCAN clustering \cite{Khan2014DBSCN} and filtering techniques \cite{kalman1960new} for obstacle motion state estimation (Section \ref{subsec:perception}), a GMM \cite{Reynolds2015Gaussian} for predicting obstacle trajectories (Section \ref{subsec:prediction}), and a planning and control module based on NeuPAN \cite{Han2025NeuPAN} (Section \ref{subsec:planning and control}). This approach leverages multi-frame data to accurately estimate obstacle motion states, enhancing the understanding of dynamic environments. It builds on NeuPAN by adding preprocessing for state estimation and trajectory forecasting, and by adjusting parameters to meet real-time demands.

\subsection{Perception}
\label{subsec:perception}
The perception module is essential for estimating the motion states of obstacles in dynamic environments. It processes a point cloud from a MID360 lidar sensor and integrates Gaussian filtering, DBSCAN clustering \cite{Khan2014DBSCN}, nearest neighbor matching, and exponential smoothing to provide reliable obstacle information.

First, the raw 3D point cloud data is downsampled and projected onto a 2D plane. This step reduces computational complexity while retaining essential spatial information. Points outside the robot's operational height range are filtered out to focus on relevant data.

Next, Gaussian filtering is applied to the projected point cloud data. This process helps to smooth the data and reduce noise, improving the quality of the point cloud for subsequent processing steps.

After filtering, DBSCAN clustering is used to identify distinct obstacles based on point density. Let \( P \) denote the filtered point cloud. For a point \(\mathbf{p} \in P\), the neighborhood \(N_{\epsilon}(\mathbf{p})\) is defined as the set of points within a distance \(\epsilon\) from \(\mathbf{p}\). A point is considered a core point if it has at least \(\textit{MinPts}\) neighbors within this distance. This step is crucial for grouping nearby points into clusters, which represent potential obstacles in the environment:
\begin{equation}
N_{\epsilon}(\mathbf{p}) = \{\mathbf{q} \in P \mid \|\mathbf{p} - \mathbf{q}\|_2 \leq \epsilon\}
\end{equation}
where \(N_{\epsilon}(\mathbf{p})\) represents the neighborhood of point \(\mathbf{p}\), \(\epsilon=1\) is the radius of the neighborhood, and \(\|\mathbf{p} - \mathbf{q}\|_2\) denotes the Euclidean distance between points \(\mathbf{p}\) and \(\mathbf{q}\).

To track obstacles across frames, the nearest neighbor matching technique is employed. For each cluster in the current frame, the nearest cluster in the previous frame is identified and matched. This matching process helps to maintain continuity in tracking obstacles over time.

Then, the motion states of obstacles are estimated using Kalman filtering. The Kalman filter recursively updates the state estimates based on new measurements, providing a robust method for tracking the velocity and position of obstacles over time.

Finally, exponential smoothing is applied to stabilize the velocity estimates. Specifically, the velocity of each cluster is estimated based on the center of the cluster:
\begin{equation}
\hat{\textbf{v}}_k = \alpha \hat{\textbf{v}}_{k-1} + (1 - \alpha) \textbf{v}_k
\end{equation}
where \(\alpha=0.7\) is the smoothing factor, \(\hat{\textbf{v}}_k\) is the updated smoothed velocity, \(\textbf{v}_k\) is current velocity measurement from the filter, and \(\hat{\textbf{v}}_{k-1}\) is previous smoothed velocity. This step ensures that the velocity estimates are smooth and reliable.

In summary, the perception module integrates Gaussian filtering, DBSCAN clustering, nearest neighbor matching, and exponential smoothing to accurately estimate the motion states of obstacles in dynamic environments, providing a robust foundation for real-time navigation and obstacle avoidance.

\subsection{Prediction}
\label{subsec:prediction}
Based on the tracked dynamic obstacles, we further predict their future positions and represent them probabilistically as a multi-frame constraint. This is achieved through a GMM \cite{Reynolds2015Gaussian} that captures the distribution of possible future states based on historical data.

The GMM is chosen for its multi-modal representation capability, computational efficiency, and interpretability. First, GMMs can effectively capture the complex and non-linear movement patterns of dynamic obstacles by modeling the distribution of possible future states as a mixture of Gaussian distributions. This multi-modal representation allows the prediction module to generate a set of potential future positions for each obstacle, reflecting the inherent uncertainty in their motion and enabling the robot to anticipate and avoid potential collisions more effectively. Second, GMMs are computationally efficient, making them suitable for real-time applications where fast and accurate predictions are crucial. Finally, GMMs offer a high degree of interpretability, as their parameters can be directly related to the physical properties of the obstacles' motion. This interpretability enhances the transparency and reliability of the prediction model, which is important for safety-critical applications.

A GMM is used to generate scattered points around obstacles. The GMM is defined as:
\begin{equation}
p(o) = \sum_{j=1}^{3} \pi_j \mathcal{N}(o | \mathbf{\mu}_j, \mathbf{\Sigma}_j)
\end{equation}

The parameters were determined through experimental tuning to best fit the specific conditions of each environment. In the simulation, the GMM parameters are set with means $[0.0, 0.1, -0.1]$, covariances $[0.002, 0.002, 0.002]$, and weights $[0.4, 0.3, 0.3]$. In the real-world experiments, the GMM parameters are means $[0.0, 0.05, 0.1]$, covariances $[0.01, 0.01, 0.01]$, and weights $[0.3, 0.5, 0.2]$. The parameters difference is due to the varying characteristics of dynamic obstacles and sensor noise in different environments.

For each obstacle point with position \(\mathbf{p}_t\) and velocity \(\mathbf{v}_t\), if the speed \(\|\mathbf{v}_t\|\) exceeds a threshold (0.3), the scattered points are generated as:
\begin{equation}
\mathbf{p}_{\text{scattered}} = \mathbf{p}_t + \mathbf{d} \cdot j \cdot \Delta s + \mathbf{p}_{\text{perp}} \cdot o
\end{equation}
where \(\mathbf{d} = \frac{\mathbf{v}_t}{\|\mathbf{v}_t\|}\) is the normalized direction vector, \(\mathbf{p}_{\text{perp}} = [-\mathbf{d}_y, \mathbf{d}_x]\) is the perpendicular vector, \(o\) is sampled from the GMM, \(j\) is the step index from 1 to \(N=20\), and \(\Delta s\) is the step size.

The prediction results are formulated as virtual obstacles (i.e., points) for motion constraint estimation within the planning module. The predicted multi-frame observation allows the robot to proactively plan its path by anticipating the future positions of dynamic obstacles, thereby improving navigation safety and efficiency.

\subsection{Planner and Controller}
\label{subsec:planning and control}

In the planner and controller module, we employ an enhanced NeuPAN framework \cite{Han2025NeuPAN} to achieve real-time path planning and obstacle avoidance. The inputs to the module include dynamic point cloud data, which consists of the current frame and virtual obstacles based on prediction. Additionally, the module takes five key waypoints from the global planner and the robot's current state, including its pose, velocity, and kinematic constraints. These inputs are transformed into a tensor \( P \in \mathbb{R}^{M \times 3} \) in the robot's coordinate system.

The core principle of the enhanced NeuPAN framework involves two main stages. First, the dynamic point cloud encoding (DUNE) stage converts the point cloud data into latent distance features (LDF), which capture the spatial relationships between the robot and obstacles. This stage introduces a velocity weight factor for dynamic obstacles, enhancing the system's sensitivity to high-speed moving objects. Second, the neural regularized motion planning (NRMP) stage optimizes the control sequence to generate robust control commands. The optimization objective in the NRMP stage is formulated as:
\begin{equation}
\min_{\mathbf{u}_{0:H}} \sum_{k=0}^{H} \| \mathbf{s}_k - \mathbf{s}_k^{\text{ref}} \|^2_{\mathbf{Q}} + \sum_{k=0}^{H} \| \mathbf{u}_k \|^2_{\mathbf{R}} + \rho_1 \sum_{i,j} \eta_j \| \text{dist}(\mathbf{s}_k, \mathbf{p}_j) \|^2
\end{equation}
where \(\mathbf{u}_{0:H}\) denotes the control sequence over a time horizon of \(H\) steps, \(\mathbf{s}_k\) represents the predicted state obtained recursively from the robot's kinematic model, \(\mathbf{Q}\) and \(\mathbf{R}\) are the weight matrices for tracking error and control effort, respectively, and \(\rho_1\) is the dynamic obstacle penalty coefficient coupled with the velocity weight \(\eta_j\).

The control command generation stage solves the optimization problem in real time and generates the control input $\mathbf{u} = [v^*, \omega^*]$. The linear velocity $v^*$ and angular velocity $\omega^*$ generated by NRMP are then processed to produce the final control commands. Specifically, the linear velocity $v^*$ is smoothed using a first-order low-pass filter to prevent abrupt changes, resulting in the smoothed linear velocity $v_{\text{out}}$. The angular velocity $\omega^*$ is clipped to ensure motor safety, resulting in the clipped angular velocity $\omega_{\text{out}}$. The key outputs of the module are the smoothed linear velocity $v_{\text{out}}$ and the clipped angular velocity $\omega_{\text{out}}$, which drive the robot and enable efficient navigation and obstacle avoidance in dynamic environments.

\section{Experiments}
\label{sec:experiments}
\subsection{Simulation Experiment}

To evaluate the effectiveness of the proposed method, systematic simulation experiments were conducted using the IR-SIM platform \cite{Han2025NeuPAN}. Given the lack of suitable open datasets for our task, we designed test cases within IR-SIM. IR-SIM is a versatile and accurate simulation environment tailored for robot navigation in complex dynamic scenarios, supporting various robot models. It serves as an ideal platform for testing and validating advanced navigation algorithms.

\begin{figure}[htbp]
    \centering
    \begin{subfigure}[b]{0.32\linewidth}
        \centering
        \includegraphics[width=\linewidth]{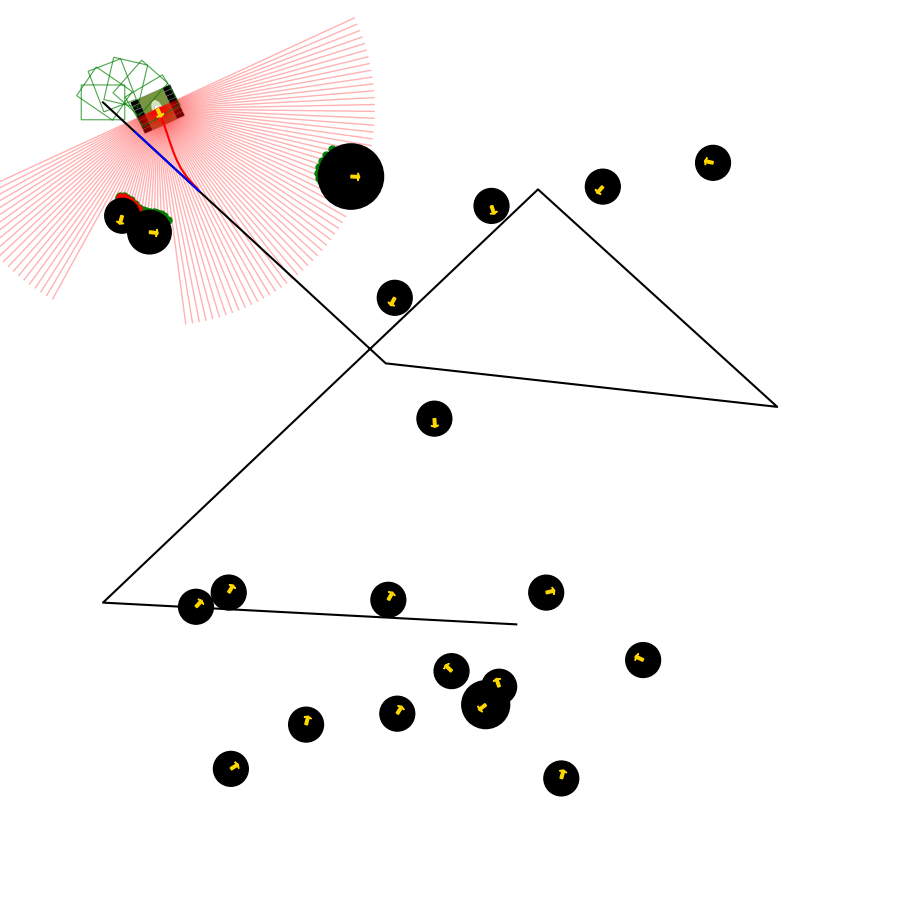}
        \label{fig:scenerio1}
    \end{subfigure}
    \begin{subfigure}[b]{0.32\linewidth}
        \centering
        \includegraphics[width=\linewidth]{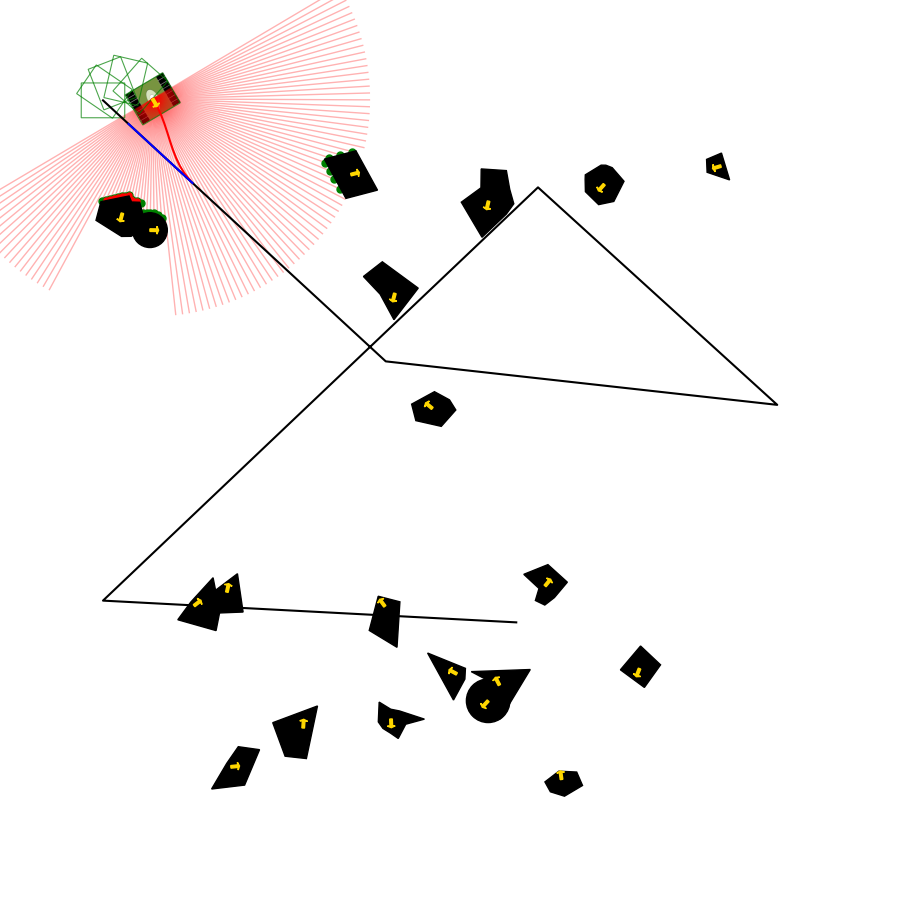}
        \label{fig:scenerio2}
    \end{subfigure}
    \begin{subfigure}[b]{0.32\linewidth}
        \centering
        \includegraphics[width=\linewidth]{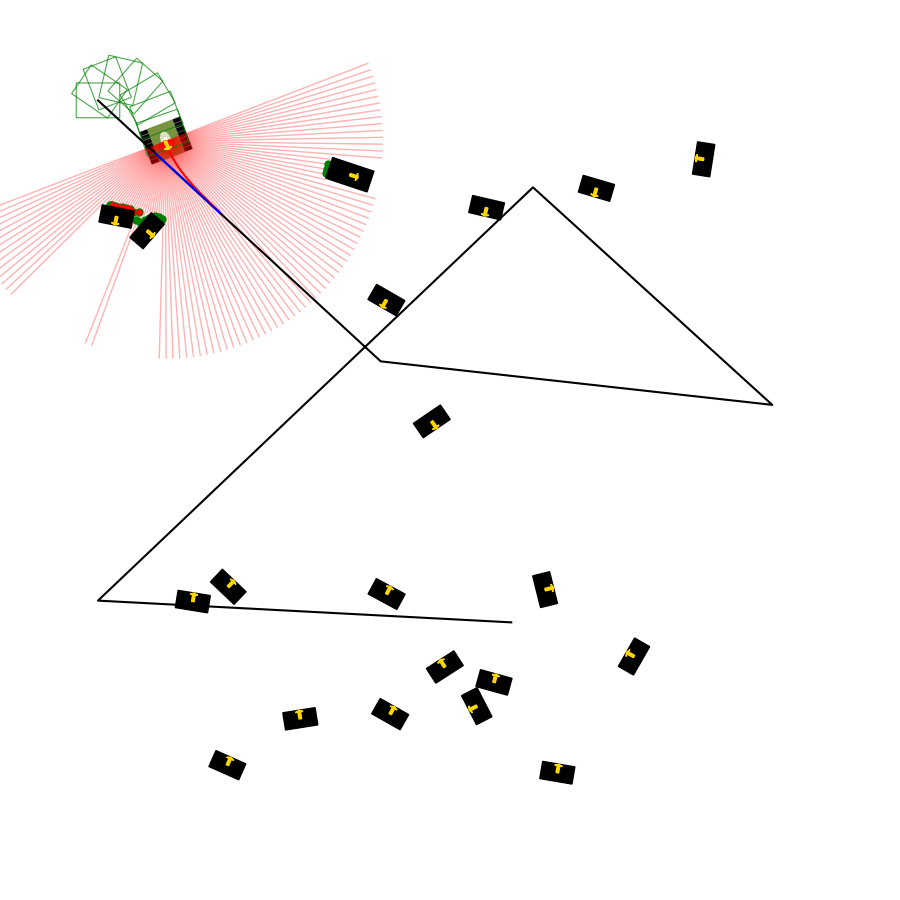}
        \label{fig:scenerio3}
    \end{subfigure}
    \caption{Various Simulation environment: The red radial lines simulate radar data for obstacle detection. The solid red line is the local planner's path, tracking the black route while avoiding obstacles. Black shapes are obstacles, with yellow arrows showing motion direction.}
    \label{fig:simulation}
\end{figure}

We tested the method in environments with high-density static and dynamic obstacles, as shown in Figure \ref{fig:simulation}. The robot, equipped with a lidar sensor, navigated from a start to a goal point. Since current methods are generally capable of avoiding obstacles, we select path length as the key metric for comparison. The optimal path length achieved is 97.5 meters.

Our method, which incorporates multi-frame constraints, consistently achieves shorter or comparable path lengths compared to NeuPAN, a single-frame method, as shown in Table \ref{tab:frames}. Notably, our method demonstrates robustness in challenging scenarios with large obstacles and higher obstacle density, where NeuPAN often gets stuck and even fails to navigate.

\begin{table}[htbp]
    \centering
    \begin{tabular}{ccc}
        \toprule
        Test Case & NeuPAN & Ours-MfNeuPAN \\ \midrule
        1 & 102.01 m & 102.00 m \\
        2 & 110.47 m & 106.28 m \\
        3 & 112.87 m & 109.35 m \\
        4 & Fail & 125.20 m \\ \bottomrule
    \end{tabular}
    \caption{Comparison of the Path Lengths between Planners Based on Single-Frame (NeuPAN) and Multi-Frame (MfNeuPAN)}
    \label{tab:frames}
\end{table}

To demonstrate the efficacy of GMM-based prediction, we compare path planning results with no prediction, a constant velocity model, and a GMM model. The results show that the GMM model achieves significantly shorter path lengths, especially in challenging scenarios.

Table \ref{tab:gmm} illustrates that the GMM model outperforms the other methods in both test cases, highlighting its effectiveness in predicting obstacle trajectories.

\begin{table}[htbp]
    \centering
    \begin{tabular}{cccc}
        \toprule
        Test Case & No Prediction & Constant Velocity Model & GMM Model \\ \midrule
        1 & 104.37 m & 103.22 m & 102.38 m \\
        2 & 115.59 m & 109.18 m & 107.92 m \\ \bottomrule
    \end{tabular}
    \caption{Comparison of the Path Lengths between Planners Based on Different Prediction Methods}
    \label{tab:gmm}
\end{table}

In addition to evaluating the performance of our planner, we also test the path length against other local planners. Specifically, we compare our method with the DWA \cite{Fiorini1998Motion}, RDA-planner \cite{Han2023RDA}, and NeuPAN \cite{Han2025NeuPAN}.

Table \ref{tab:other} presents the results, demonstrating that our method outperforms other methods. That suggests our planner is well-suited for applications where minimizing path length is crucial, and it can serve as a reliable alternative to existing methods. At the same time, compared with state-of-the-art methods like DRL-VO, we can deal with point constraints, which are not specifically designed for people obstacles, showing more prospects in the real world.

\begin{table}[htbp]
    \centering
    \begin{tabular}{ccccc}
        \toprule
        Test Case & DWA & RDA Planner & NeuPAN & MfNeuPAN \\ \midrule
        1 & 103.42 m & 103.22 m & 106.55 m & 103.11 m \\
        2 & 111.00 m & Fail & 104.37 m & 102.38 m \\ \bottomrule
    \end{tabular}
    \caption{Comparison of the Path Lengths between Different Planners}
    \label{tab:other}
\end{table}

The average time required to process an observation result into a control sequence is 62.60 ms, which is sufficient to meet practical application requirements. The experiments were conducted on a standard desktop computer equipped with an Intel Core i7 CPU, 16 GB of RAM, and an NVIDIA GeForce RTX 2060 GPU.

\subsection{Real-World Experiment}

To further validate the applicability and robustness of the proposed method in real-world scenarios, experiments were conducted in a laboratory environment. The laboratory was chosen as it includes both static and dynamic obstacles, providing a controlled yet complex setting to test the navigation and obstacle avoidance capabilities of the robot.

The experimental setup included a Scout robot equipped with a MID360 lidar sensor. The robot was tasked with navigating along a predefined path provided by the odometry of FAST-LIO \cite{Xu2021Fast}, while avoiding various types of obstacles, including both static and dynamic objects. The predefined path was designed to simulate a typical indoor navigation scenario, with obstacles placed to create narrow passages and dynamic interactions.

\begin{figure}[htbp]
    \centering
    \begin{subfigure}[b]{0.49\linewidth}
        \centering
        \includegraphics[width=\linewidth]{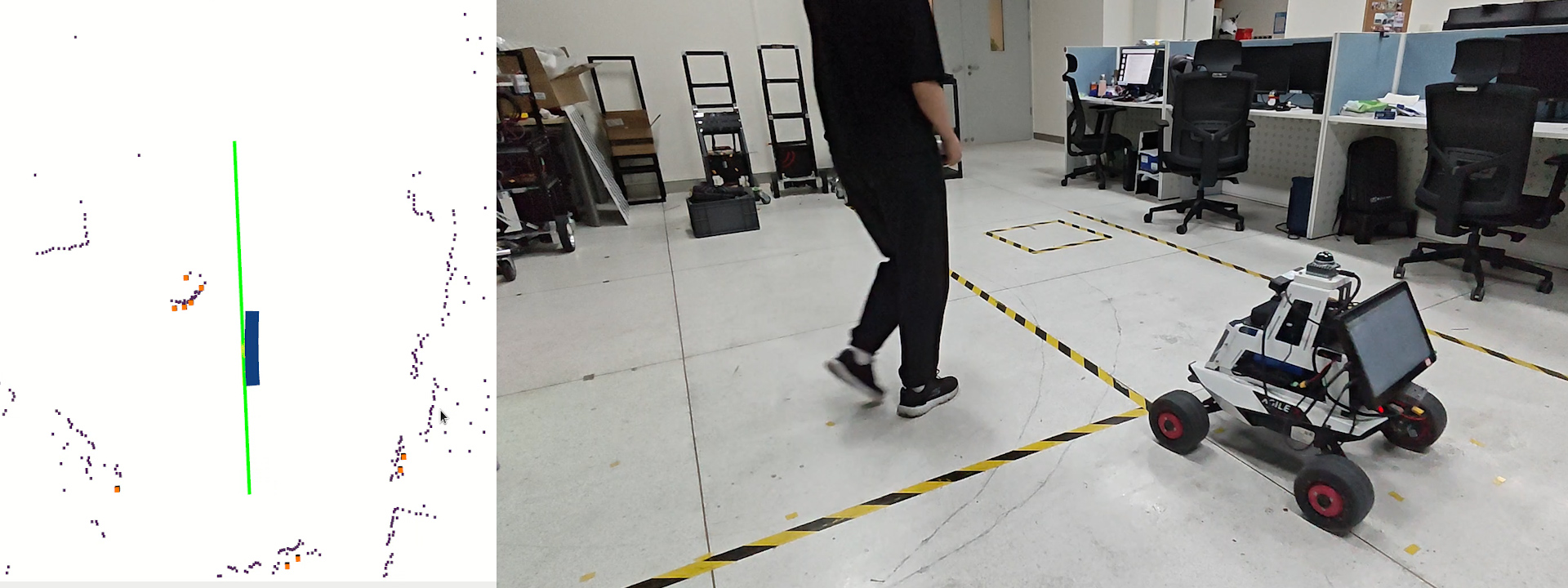}
        \label{fig:fail1}
    \end{subfigure}
    \begin{subfigure}[b]{0.49\linewidth}
        \centering
        \includegraphics[width=\linewidth]{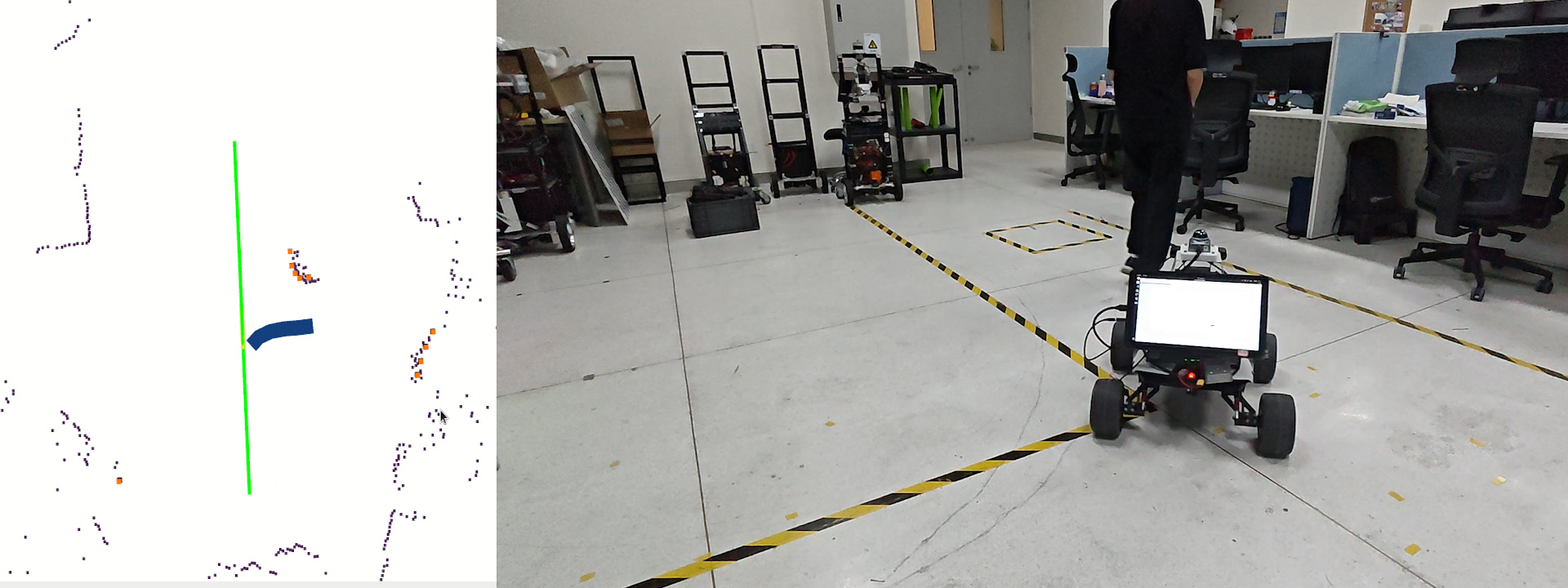}
        \label{fig:fail2}
    \end{subfigure}
    \caption{NeuPAN's reactive obstacle avoidance strategy. In frame n-1 (left), the robot encounters an obstacle. By frame n (right), it reacts by taking a right detour to avoid collision.}
    \label{fig:failures}
\end{figure}

\begin{figure}[htbp]
    \centering
    \begin{subfigure}[b]{0.49\linewidth}
        \centering
        \includegraphics[width=\linewidth]{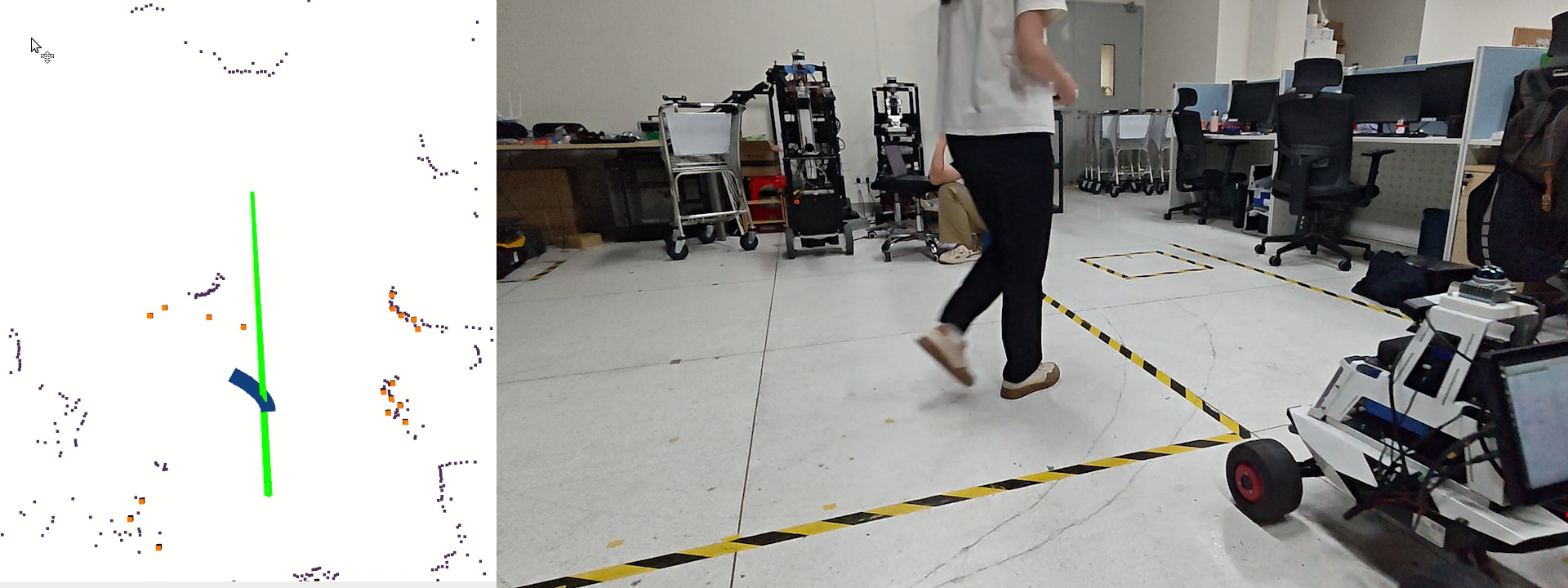}
        \label{fig:success1}
    \end{subfigure}
    \begin{subfigure}[b]{0.49\linewidth}
        \centering
        \includegraphics[width=\linewidth]{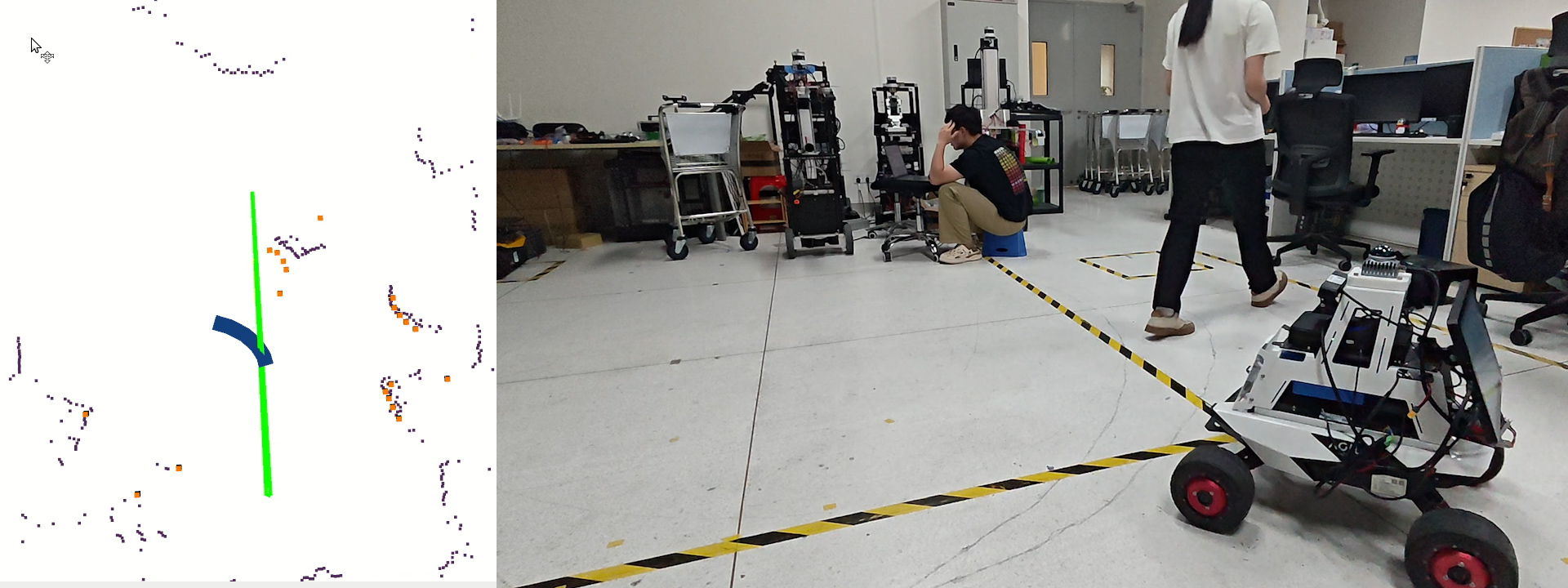}
        \label{fig:success2}
    \end{subfigure}
    \caption{MfNeuPAN's proactive obstacle avoidance strategy. In frame n-1 (left), the robot anticipates an obstacle on the right. By frame n (right), it proactively turns left to avoid the obstacle.}
    \label{fig:successes}
\end{figure}

These results highlight the effectiveness of our multi-frame-based proactive planning approach in enabling the robot to dynamically adjust its path in real-time and successfully navigate through complex dynamic environments. As shown in Figure \ref{fig:failures}, the baseline often fails to avoid dynamic obstacles effectively, thus continuing to take detours, or even leading to collisions or getting stuck. In contrast, our method, as illustrated in Figure \ref{fig:successes}, successfully navigates the robot to avoid dynamic obstacles by taking appropriate detours. The complete experimental results are in the supplementary video. The robot's ability to dynamically adjust its path in real-time is a significant advantage in complex environments. In our 20 experimental trials, the robot successfully avoided all obstacles without any collisions.

\section{Conclusion}
\label{sec:conclusion}
This paper introduces a proactive end-to-end navigation framework for robots in dynamic environments, utilizing multi-frame point constraints to significantly enhance real-time navigation and obstacle avoidance. By integrating perception, prediction, planning, and control modules, our framework effectively tackles the challenges of dynamic obstacles, sensor noise, and constrained spaces. The key innovation lies in the use of multi-frame observations, which include both current and predicted future frames. These observations allow the robot to proactively anticipate and avoid dynamic obstacles. This is enabled by combining DBSCAN clustering for obstacle detection and tracking, Gaussian Mixture Model for representing the predicted trajectories of obstacles, and an enhanced NeuPAN method for robust planning and control. Both simulation and real-world experiments confirm the superior performance and robustness of our framework, showcasing its potential to revolutionize end-to-end navigation in complex dynamic environments.

\bibliographystyle{IEEEtran}
\bibliography{main}

@inbook{Reynolds2015Gaussian,
  author = {D. Reynolds},
  title = {Gaussian mixture models},
  booktitle = {Encyclopedia of Biometrics},
  publisher = {Springer},
  address = {Boston, MA},
  pages = {827--832},
  year = {2015}
}

@article{Xu2021Fast,
  author = {W. Xu and F. Zhang},
  title = {Fast-lio: A fast, robust lidar-inertial odometry package by tightly-coupled iterated kalman filter},
  journal = {IEEE Robotics and Automation Letters},
  volume = {6},
  number = {2},
  pages = {3317--3324},
  year = {2021}
}

@inproceedings{Khan2014DBSCN,
  author = {Khan, K. and Rehman, S. U. and Aziz, K. and others},
  title = {DBSCAN: Past, present and future},
  booktitle = {The fifth international conference on the applications of digital information and web technologies (ICADIWT 2014)},
  pages = {232--238},
  year = {2014},
  organization = {IEEE}
}

@article{Falanga2020Dynamic,
  author = {D. Falanga and K. Kleber and D. Scaramuzza},
  title = {Dynamic obstacle avoidance for quadrotors with event cameras},
  journal = {Science Robotics},
  volume = {5},
  number = {40},
  year = {2020}
}

@inproceedings{Lin2020Robust,
  author = {J. Lin and H. Zhu and J. Alonso-Mora},
  title = {Robust vision-based obstacle avoidance for micro aerial vehicles in dynamic environments},
  booktitle = {Proc. IEEE ICRA},
  pages = {2682--2688},
  year = {2020}
}

@inproceedings{Pokle2019Deep,
  author = {A. Pokle and others},
  title = {Deep local trajectory replanning and control for robot navigation},
  booktitle = {Proc. IEEE ICRA},
  pages = {5815--5822},
  year = {2019}
}

@inproceedings{Ester1996Density,
  author = {M. Ester and others},
  title = {A density-based algorithm for discovering clusters in large spatial databases with noise},
  booktitle = {Proc. KDD},
  pages = {226--231},
  year = {1996}
}

@article{Chen2022Real,
  author = {H. Chen and P. Lu},
  title = {Real-time identification and avoidance of simultaneous static and dynamic obstacles on point cloud for UAVs navigation},
  journal = {Robotics and Autonomous Systems},
  volume = {154},
  year = {2022}
}

@inproceedings{Arpino2021Robot,
  author = {C. P\'{e}rez-D'Arpino and others},
  title = {Robot navigation in constrained pedestrian environments using reinforcement learning},
  booktitle = {Proc. IEEE ICRA},
  pages = {1140--1146},
  year = {2021}
}

@article{Mersch2022Receding,
  author = {B. Mersch and others},
  title = {Receding moving object segmentation in 3D {LiDAR} data using sparse 4D convolutions},
  journal = {IEEE Robot. Autom. Lett.},
  volume = {7},
  number = {3},
  pages = {7503--7510},
  year = {2022}
}

@inproceedings{Tai2016ADeep,
  author = {L. Tai and S. Li and M. Liu},
  title = {A deep-network solution towards model-less obstacle avoidance},
  booktitle = {Proc. IEEE IROS},
  pages = {2759--2764},
  year = {2016}
}

@article{Lu2024FAPP,
  author = {M. Lu and others},
  title = {{FAPP}: Fast and Adaptive Perception and Planning for {UAVs} in Dynamic Cluttered Environments},
  journal = {IEEE Trans. Robot.},
  year = {2024}
}

@article{Fiorini1998Motion,
  author = {P. Fiorini and Z. Shiller},
  title = {Motion planning in dynamic environments using velocity obstacles},
  journal = {Int. J. Robot. Res.},
  volume = {17},
  number = {7},
  pages = {760--772},
  year = {1998}
}

@inproceedings{VanDenBerg2011Reciprocal,
  author = {J. Van Den Berg and others},
  title = {Reciprocal N-body collision avoidance},
  booktitle = {Robotics Research},
  pages = {3--19},
  year = {2011}
}

@article{Brito2019Model,
  author = {B. Brito and others},
  title = {Model predictive contouring control for collision avoidance in unstructured dynamic environments},
  journal = {IEEE Robot. Autom. Lett.},
  volume = {4},
  number = {4},
  pages = {4459--4466},
  year = {2019}
}

@inproceedings{Xu2022DPMPC,
  author = {Z. Xu and others},
  title = {{DPMPC}-Planner: A real-time {UAV} trajectory planning framework for complex static environments with dynamic obstacles},
  booktitle = {Proc. IEEE ICRA},
  pages = {250--256},
  year = {2022}
}

@article{Xie2023DRL,
  author = {Z. Xie and P. Dames},
  title = {{DRL-VO}: Learning to navigate through crowded dynamic scenes using velocity obstacles},
  journal = {IEEE Trans. Robot.},
  volume = {39},
  number = {4},
  pages = {2700--2719},
  year = {2023}
}

@article{Xu2025NavRL,
  author = {Z. Xu and others},
  title = {{NavRL}: Learning Safe Flight in Dynamic Environments},
  journal = {IEEE Robot. Autom. Lett.},
  year = {2025}
}

@inproceedings{Liu2020Robot,
  author = {L. Liu and others},
  title = {Robot navigation in crowded environments using deep reinforcement learning},
  booktitle = {Proc. IEEE IROS},
  year = {2020}
}

@article{Han2025NeuPAN,
  author = {R. Han and others},
  title = {{NeuPAN}: Direct Point Robot Navigation with End-to-End Model-Based Learning},
  journal = {IEEE Trans. Robot.},
  year = {2025}
}

@article{Zhang2020Optimization,
  author = {X. Zhang and A. Liniger and F. Borrelli},
  title = {Optimization-based collision avoidance},
  journal = {IEEE Transactions on Control Systems Technology},
  volume = {29},
  number = {3},
  pages = {972--983},
  year = {2020}
}

@inproceedings{Rosmann2017Kinodynamic,
  author = {C. Rosmann and F. Hoffmann and T. Bertram},
  title = {Kinodynamic trajectory optimization and control for car-like robots},
  booktitle = {2017 IEEE/RSJ International Conference on Intelligent Robots and Systems (IROS)},
  pages = {5681--5686},
  year = {2017}
}

@article{Salzman2016Asymptotically,
  author = {O. Salzman and D. Halperin},
  title = {Asymptotically near-optimal RRT for fast, high-quality motion planning},
  journal = {IEEE Transactions on Robotics},
  volume = {32},
  number = {3},
  pages = {473--483},
  year = {2016}
}

@article{Han2023RDA,
  author = {R. Han and S. Wang and S. Wang and Z. Zhang and Q. Zhang and Y. C. Eldar and Q. Hao and J. Pan},
  title = {{RDA}: An accelerated collision-free motion planner for autonomous navigation in cluttered environments},
  journal = {IEEE Robotics and Automation Letters},
  volume = {8},
  number = {3},
  pages = {1715--1722},
  year = {2023}
}

@article{Devo2020Towards,
  author = {A. Devo and G. Mezzetti and G. Costante and M. L. Fravolini and P. Valigi},
  title = {Towards generalization in target-driven visual navigation by using deep reinforcement learning},
  journal = {IEEE Transactions on Robotics},
  volume = {36},
  number = {5},
  pages = {1546--1551},
  year = {2020}
}

@article{Fan2020Distributed,
  author = {T. Fan and P. Long and W. Liu and J. Pan},
  title = {Distributed multi-robot collision avoidance via deep reinforcement learning for navigation in complex scenarios},
  journal = {The International Journal of Robotics Research},
  volume = {39},
  number = {7},
  pages = {856--892},
  year = {2020}
}

@article{Qureshi2020Motion,
  author = {A. H. Qureshi and Y. Miao and A. Simeonov and M. C. Yip},
  title = {Motion planning networks: Bridging the gap between learning-based and classical motion planners},
  journal = {IEEE Transactions on Robotics},
  volume = {37},
  number = {1},
  pages = {48--66},
  year = {2020}
}

@inproceedings{Wang2022Adaptive,
  author = {S. Wang and R. Gao and R. Han and S. Chen and C. Li and Q. Hao},
  title = {Adaptive environment modeling based reinforcement learning for collision avoidance in complex scenes},
  booktitle = {2022 IEEE/RSJ International Conference on Intelligent Robots and Systems (IROS)},
  pages = {9011--9018},
  year = {2022}
}

@article{Li2021MPC,
  author = {L. Li and Y. Miao and A. H. Qureshi and M. C. Yip},
  title = {{MPC-MPNet}: Model-predictive motion planning networks for fast, near-optimal planning under kinodynamic constraints},
  journal = {IEEE Robotics and Automation Letters},
  volume = {6},
  number = {3},
  pages = {4496--4503},
  year = {2021}
}

@article{Fox1997Dynamic,
  author = {D. Fox and W. Burgard and S. Thrun},
  title = {The dynamic window approach to collision avoidance},
  journal = {IEEE Robotics \& Automation Magazine},
  volume = {4},
  number = {1},
  pages = {23--33},
  year = {1997}
}

@inproceedings{Quigley2009ROS,
  author = {M. Quigley and K. Conley and B. Gerkey and J. Faust and T. Foote and J. Leibs and R. Wheeler and A. Y. Ng},
  title = {{ROS}: an open-source robot operating system},
  booktitle = {ICRA Workshop on Open Source Software},
  volume = {3},
  number = {3.2},
  address = {Kobe, Japan},
  pages = {5},
  year = {2009}
}

@article{GARCIA1989335,
  author = {Carlos E. García and David M. Prett and Manfred Morari},
  title = {Model predictive control: Theory and practice—A survey},
  journal = {Automatica},
  volume = {25},
  number = {3},
  pages = {335--348},
  year = {1989}
}

@article{kalman1960new,
  title={A new approach to linear filtering and prediction problems},
  author={Kalman, Rudolph Emil},
  year={1960}
}

@inproceedings{redmon2016you,
  title={You only look once: Unified, real-time object detection},
  author={Redmon, Joseph and Divvala, Santosh and Girshick, Ross and Farhadi, Ali},
  booktitle={Proceedings of the IEEE conference on computer vision and pattern recognition},
  pages={779--788},
  year={2016}
}

\end{document}